\documentclass[11pt,a4paper]{article}
\usepackage[hyperref]{naaclhlt2019}
\usepackage{times}
\usepackage{latexsym}
\usepackage{amsmath}
\usepackage{booktabs}

\usepackage{url}

\aclfinalcopy 

\title{Acquisition of Inflectional Morphology in Artificial Neural Networks\\With Prior Knowledge}

\author{  
  Katharina Kann\\ 
  New York University, USA\\
  \texttt{kann@nyu.edu} 
}

\date{}

\begin{document}
\maketitle
\begin{abstract}
  How does knowledge of one language's morphology influence learning of inflection rules in a second one? In order to investigate this question in artificial neural network models, we perform experiments with a sequence-to-sequence architecture, which we train on different combinations of eight source and three target languages. A detailed analysis of the model outputs suggests the following conclusions: (i) if source and target language are closely related, acquisition of the target language's inflectional morphology constitutes an easier task for the model; (ii) knowledge of a prefixing (resp. suffixing) language makes acquisition of a suffixing (resp. prefixing) language's morphology more challenging; and (iii) surprisingly, a source language which exhibits an agglutinative morphology simplifies learning of a second language's inflectional morphology, independent of their relatedness.
\end{abstract}

\section{Introduction}
A widely agreed-on fact in language acquisition research is that learning of a second language (L2) is influenced by a learner's native language (L1) \cite{dulay1974natural,kellerman1979transfer}. A language's morphosyntax seems to be no exception to this rule \cite{bliss2006l2}, but the exact nature of this influence remains unknown. For instance, it is unclear whether it is constraints imposed by the phonological or by the morphosyntactic attributes of the L1 that are more important during the process of learning an L2's morphosyntax.

 \begin{table}[t] 
  \setlength{\tabcolsep}{7.5pt}
  \centering
  \begin{tabular}{l @{\hspace*{.8cm}} l l}
    \toprule
    & \textbf{walk} & \textbf{eat} \\
    \midrule
    \texttt{Inf} & {dance} & { eat} \\
    \texttt{3rdSgPres} & {dances} & { eats} \\
    \texttt{PresPart} & {dancing} & { eating} \\
    \texttt{Past} & {danced} & { ate} \\
    \texttt{PastPart} & {danced} & { eaten} \\
    \bottomrule
  \end{tabular}
  \caption{\label{tab:paradigms} Paradigms of the English lemmas \textit{dance} and \textit{eat}. \textit{dance} has $4$ distinct inflected forms; \textit{eat} has $5$.
  } 
\end{table}
Within the area of natural language processing (NLP) research, experimenting on neural network models just as if they were human subjects  has recently been gaining popularity \cite{ettinger-etal-2016-probing,ettinger-etal-2017-towards,kim-etal-2019-probing}. Often, so-called probing tasks are used, which require a specific subset of linguistic knowledge and can, thus, be leveraged for qualitative evaluation.
The goal is to answer the question: What do neural networks learn that helps them to succeed in a given task?

Neural network models, and specifically sequence-to-sequence models, have pushed the state of the art for morphological inflection -- the task of learning a mapping from lemmata to their inflected forms -- in the last years \cite{W16-2002}.
Thus, in this work, we experiment on such models, asking not \textit{what} they learn, but, motivated by the respective research on human subjects, the related
question of \textit{how what they learn depends on their prior knowledge}. We manually investigate the errors made by artificial neural networks for morphological inflection in a target language after pretraining on different source languages.
We aim at finding answers to two main questions: (i) Do errors systematically differ between source languages? (ii) Do these differences seem explainable, given the properties of the source and target languages?
In other words, we are interested in exploring if and how L2 acquisition of morphological inflection depends on the L1, i.e., the "native language", in neural network models.

To this goal, we select a diverse set of eight source languages from different language families -- Basque, French, German, Hungarian, Italian, Navajo, Turkish, and Quechua -- and three target languages -- English, Spanish and Zulu. We pretrain a neural sequence-to-sequence architecture on each of the source languages and then fine-tune the resulting models on small datasets in each of the target languages. Analyzing the errors made by the systems, we find that (i) source and target language being closely related simplifies the successful learning of inflection in the target language, (ii) the task is harder to learn in a prefixing language if the source language is suffixing -- as well as the other way around, and (iii) a source language which exhibits an agglutinative morphology simplifies learning of a second language's inflectional morphology.

\section{Task}
Many of the world's languages exhibit rich inflectional morphology: the surface form of an individual lexical entry changes in order to express properties such as person, grammatical gender, or case. The citation form of a lexical entry is referred to as the \textit{lemma}. The set of all possible surface forms or \textit{inflections} of a lemma is called its \textit{paradigm}. 
Each inflection within a paradigm can be associated with a tag, i.e., \texttt{3rdSgPres} is the morphological tag associated with the inflection \textit{dances} of the English lemma \textit{dance}.
We display the paradigms of \textit{dance} and \textit{eat} in Table \ref{tab:paradigms}.

The presence of rich inflectional morphology is problematic for NLP systems as
it increases word form sparsity. For instance,
while English verbs can have up to $5$ inflected forms, 
Archi verbs
have thousands \cite{kibrik2017archi}, even by a conservative count.
Thus, an important task in the area of morphology is morphological inflection \cite{durrett-denero:2013:NAACL-HLT,cotterell-EtAl:2018:K18-30}, which consists of mapping a lemma to an indicated inflected form.
An (irregular) English example would be
\begin{align*}
(\textrm{eat},  \textrm{\texttt{PAST}}) \rightarrow \textrm{ate}
\end{align*}
with 
\texttt{PAST} being the target tag, denoting the past tense form.
Additionally, a rich inflectional morphology is also challenging for L2 language learners, since both rules and their exceptions need to be memorized.

In NLP, morphological inflection has recently frequently been cast as a sequence-to-sequence problem, where the sequence of target (sub-)tags together with the sequence of input characters constitute the input sequence, and the characters of the inflected word form the output. Neural models define the state of the art for the task and obtain high accuracy if an abundance of training data is available. 
Here, we focus on learning of inflection from limited data if information about another language's morphology is already known. We, thus, loosely simulate an L2 learning setting. 

\paragraph{Formal definition. } 
Let ${\cal M}$ be 
the paradigm slots which are being expressed in a language, and $w$ a lemma in that language. 
We then define the paradigm $\pi$ of $w$ as:
\begin{equation}
  \pi(w) = \Big\{ \big( f_k[w], t_{k} \big) \Big\}_{k \in {\cal M}(w)}
\end{equation}
$f_k[w]$ denotes an inflected form corresponding to tag $t_{k}$, and $w$ and $f_k[w]$ are strings consisting of letters from an alphabet $\Sigma$.

The task of morphological inflection consists of predicting a missing form $f_i[w]$ from a paradigm, given the lemma $w$ together with the tag $t_i$.

\section{Model}
\subsection{Pointer--Generator Network}
The models we experiment with are based on a pointer--generator network architecture \cite{gu-EtAl:2016:P16-1,P17-1099}, i.e., a recurrent neural network (RNN)-based sequence-to-sequence network with attention and a copy mechanism. A standard sequence-to-sequence model \cite{bahdanau2015neural} has been shown to perform well for morphological inflection \cite{kann-schutze:2016:P16-2} and has, thus, been subject to cognitively motivated experiments \cite{TACL1420.bib} before. Here, however, we choose the pointer--generator variant of \newcite{sharma-katrapati-sharma:2018:K18-30}, since it performs better in low-resource settings, which we will assume for our target languages.
We will explain the model shortly in the following and refer the reader to the original paper for more details.

\paragraph{Encoders. }
Our architecture employs two separate encoders, which are both bi-directional long short-term memory (LSTM) networks \cite{hochreiter1997long}: The first processes the morphological tags which describe the desired target form one by one.\footnote{In contrast to other work on cross-lingual transfer in deep learning models we do not employ language embeddings.} The second encodes the sequence of characters of the input word. 

\paragraph{Attention. }
Two separate attention mechanisms are used: one per encoder LSTM. Taking all respective encoder hidden states as well as the current decoder hidden state as input, each of them outputs a so-called context vector, which is a weighted sum of all encoder hidden states. The concatenation of the two individual context vectors results in the final context vector $c_t$, which is the input to the decoder at time step $t$.

\paragraph{Decoder. }
Our decoder consists of a uni-directional LSTM. Unlike a standard sequence-to-sequence model, a pointer--generator network is not limited to generating characters from the vocabulary to produce the output.
Instead, the model gives certain probability to copying elements from the input over to the output.
The probability of a character $y_t$ at time step $t$ is
computed as a sum of the probability of $y_t$ given by the decoder and the probability of copying $y_t$, weighted by the probabilities of generating and copying:
\begin{equation}
    p(y_t) = \alpha p_{\textrm{dec}}(y_t) + (1-\alpha) p_{\textrm{copy}}(y_t)
\end{equation}
$p_{\textrm{dec}}(y_t)$ is calculated as an LSTM update and a projection of the decoder state to the vocabulary, followed by a softmax function. $p_{\textrm{copy}}(y_t)$ corresponds to the attention weights for each input character.
The model computes the probability $\alpha$ with which it generates a new output character as
\begin{equation}
\alpha = \sigma(w_c c_t + w_s s_t + w_y y_{t-1} + b)
\end{equation}
for context vector $c_t$, decoder state $s_t$, embedding of the last output $y_{t-1}$, weights $w_c$, $w_s$, $w_y$, and bias vector $b$.
It has been shown empirically that the copy mechanism of the pointer--generator network architecture is beneficial for morphological generation in the low-resource setting \cite{sharma-katrapati-sharma:2018:K18-30}.

\subsection{Pretraining and Finetuning}
Pretraining and successive fine-tuning of neural network models is a common approach for handling of low-resource settings
in NLP. The idea is that certain properties of language can be learned either from raw text, related tasks, or related languages. Technically, \textit{pretraining} consists of estimating some or all model parameters on examples which do not necessarily belong to the final target task. \textit{Fine-tuning} refers to continuing training of such a model on a target task, whose data is often limited. While the sizes of the pretrained model parameters usually remain the same between the two phases, the learning rate or other details of the training regime, e.g., dropout, might differ. Pretraining can be seen as finding a suitable initialization of model parameters, before training on limited amounts of task- or language-specific examples.

In the context of morphological generation, pretraining in combination with fine-tuning has been used by \newcite{kann-schutze-2018-neural}, which proposes to pretrain a model on general inflection data and fine-tune on examples from a specific paradigm whose remaining forms should be automatically generated. 
Famous examples for pretraining in the wider area of NLP include BERT \cite{devlin-etal-2019-bert} or GPT-2 \cite{radford2019language}: there, general properties of language are learned using large unlabeled corpora.

Here, we are interested in pretraining as a simulation of familiarity with a native language. By investigating a fine-tuned model we ask the question: How does extensive knowledge of one language influence the acquisition of another?

\section{Experimental Design}

\subsection{Target Languages}
We choose three target languages.

English (ENG) is a morphologically impoverished language, as far as inflectional morphology is concerned. Its verbal paradigm only consists of up to 5 different forms and its nominal paradigm of only up to 2. However, it is one of the most frequently spoken and taught languages in the world, making its acquisition a crucial research topic. 

Spanish (SPA), in contrast, is morphologically rich, and disposes of much larger verbal paradigms than English. Like English, it is a suffixing language, and it additionally makes use of internal stem changes (e.g., \textit{o} $\rightarrow$ \textit{ue}).

Since English and Spanish are both Indo-European languages, and, thus, relatively similar, we further add a third, unrelated target language. We choose Zulu (ZUL), a Bantoid language. In contrast to the first two, it is strongly prefixing.

\begin{table*}[t] 
  \setlength{\tabcolsep}{2.5pt}
  \small
  \centering
  \begin{tabular}{l l | l l l | l l l l l l l l}
  \toprule
& & ENG & SPA & ZUL & EUS & FRA & DEU & HUN & ITA & NAV & TUR & QVH \\\midrule
20A & Fusion of Selected Inflectional Formatives & 0 & 0 & 0 & 0 & 0 & 0 & 0 & 1 & 0 & 0 & 0 \\
21A & Exponence of Selected Inflectional Formatives & 0 & 1 & 0 & 1 & 0 & 2 & 1 & 3 & 3 & 1 & 1 \\
21B & Exponence of Tense-Aspect-Mood Inflection & 0 & 1 & 0 & 0 & 1 & 0 & 0 & 2 & 2 & 0 & 0 \\
22A & Inflectional Synthesis of the Verb & 0 & 1 & 1 & 1 & 1 & 0 & 1 & 2 & 2 & 3 & 4 \\
23A & Locus of Marking in the Clause & 0 & 1 & 2 & 1 & 3 & 0 & 0 & 4 & 4 & 0 & 0 \\
24A & Locus of Marking in Possessive Noun Phrases & 0 & 0 & 0 & 0 & 0 & 0 & 0 & 1 & 1 & 2 & 0 \\
25A & Locus of Marking: Whole-language Typology & 0 & 1 & 1 & 1 & 1 & 0 & 1 & 2 & 2 & 1 & 0 \\
25B & Zero Marking of A and P Arguments & 0 & 0 & 0 & 0 & 0 & 0 & 0 & 1 & 1 & 0 & 0 \\
26A & Prefixing vs. Suffixing in Inflectional Morphology & 0 & 0 & 1 & 2 & 0 & 0 & 0 & 0 & 1 & 0 & 0 \\
27A & Reduplication & 0 & 0 & 1 & 2 & 0 & 0 & 2 & 0 & 0 & 2 & 1 \\
28A & Case Syncretism & 0 & 1 & 2 & 0 & 1 & 1 & 3 & 4 & 2 & 3 & 3 \\
29A & Syncretism in Verbal Person/Number Marking & 0 & 0 & 0 & 1 & 0 & 0 & 1 & 2 & 1 & 1 & 1 \\
  \bottomrule
  \end{tabular}
  \caption{\label{tab:features} WALS features from the \textit{Morphology} category. 20A: 0=Exclusively concatenative, 1=N/A.
21A: 0=No case, 1=Monoexponential case, 2=Case+number, 3=N/A.
21B: 0=monoexponential TAM, 1=TAM+agreement, 2=N/A.
22A: 0=2-3 categories per word, 1=4-5 categories per word, 2=N/A, 3=6-7 categories per word, 4=8-9 categories per word.
23A: 0=Dependent marking, 1=Double marking, 2=Head marking, 3=No marking, 4=N/A.
24A: 0=Dependent marking, 1=N/A, 2=Double marking.
25A: 0=Dependent-marking, 1=Inconsistent or other, 2=N/A.
25B: 0=Non-zero marking, 1=N/A.
26A: 0=Strongly suffixing, 1=Strong prefixing, 2=Equal prefixing and suffixing.
27A: 0=No productive reduplication, 1=Full reduplication only, 2=Productive full and partial reduplication.
28A: 0=Core cases only, 1=Core and non-core, 2=No case marking, 3=No syncretism, 4=N/A.
29A: 0=Syncretic, 1=Not syncretic, 2=N/A.
} 
\end{table*}
\subsection{Source Languages}
For pretraining, we choose languages with different degrees of relatedness and varying morphological similarity to English, Spanish, and Zulu. We limit our experiments to languages which are written in Latin script.

As an estimate for morphological similarity we look at the features from the \textit{Morphology} category mentioned in The World Atlas of Language Structures (WALS).\footnote{\url{https://wals.info}} An overview of the available features as well as the respective values for our set of languages is shown in Table \ref{tab:features}.

We decide on Basque (EUS), French (FRA), German (DEU), Hungarian (HUN), Italian (ITA), Navajo (NAV), Turkish (TUR), and Quechua (QVH) as source languages.

Basque is a language isolate. Its inflectional morphology makes similarly frequent use of prefixes and suffixes, with suffixes mostly being attached to nouns, while prefixes and suffixes can both be employed for verbal inflection.

French and Italian are Romance languages, and thus belong to the same family as the target language Spanish. Both are suffixing and fusional languages. 

German, like English, belongs to the Germanic language family. It is a fusional, predominantly suffixing language and, similarly to Spanish, makes use of stem changes.

Hungarian, a Finno-Ugric language, and Turkish, a Turkic language, both exhibit an agglutinative morphology, and are predominantly suffixing. They further have vowel harmony systems.

Navajo is an Athabaskan language and the only source language which is strongly prefixing. It further exhibits consonant harmony among its sibilants \cite{rice2000morpheme,hansson2010consonant}.

Finally, Quechua, a Quechuan language spoken in South America, is again predominantly suffixing and unrelated to all of our target languages.

\subsection{Hyperparameters and Data}
We mostly use the default hyperparameters by \newcite{sharma-katrapati-sharma:2018:K18-30}.\footnote{\url{github.com/abhishek0318/conll-sigmorphon-2018}} 
In particular, all RNNs have one hidden layer of size 100, and all input and output embeddings are 300-dimensional. 

For optimization, we use ADAM \cite{kingma2014adam}. Pretraining on the source language is done for exactly 50 epochs. To obtain our final models, we then fine-tune different copies of each pretrained model for 300 additional epochs for each target language.
We employ dropout \cite{srivastava2014dropout} with a coefficient of 0.3 for pretraining and, since that dataset is smaller, with a coefficient of 0.5 for fine-tuning.

We make use of the datasets from the CoNLL--SIGMORPHON 2018 shared task \cite{cotterell-EtAl:2018:K18-30}. The organizers provided a low, medium, and high setting for each language, with 100, 1000, and 10000 examples, respectively. For all L1 languages, we train our models on the high-resource datasets with 10000 examples. For fine-tuning, we use the low-resource datasets.

\section{Quantitative Results}
\begin{table}[t] 
  \setlength{\tabcolsep}{2.5pt}
  \small
  \centering
  \begin{tabular}{l | l l l l l l l l}
  \toprule
    & EUS & FRA & DEU & HUN & ITA & NAV & TUR & QVH \\ \midrule
    ENG & 45.8 & 76.1 & 82.0 & 85.6 & 84.7 & 53.2 & 81.7 & 68.3 \\ 
 SPA & 23.9 & 53.3 & 53.8 & 58.2 & 56.9 & 33.1 & 52.0 & 49.0 \\ 
 ZUL & 10.8 & 17.1 & 23.0 & 23.0 & 21.9 & 13.6 & 24.9 & 10.7 \\
  \bottomrule
  \end{tabular}
  \caption{\label{tab:acc} Test accuracy.} 
\end{table}
In Table \ref{tab:acc}, we show the final test accuracy for all models and languages. 
Pretraining on EUS and NAV results in the weakest target language inflection models for ENG, which might be explained by  those two languages being unrelated to ENG and making at least partial use of prefixing, while ENG is a suffixing language (cf. Table \ref{tab:features}). In contrast, HUN and ITA yield the best final models for ENG. This is surprising, since DEU is the language in our experiments which is closest related to ENG. 

For SPA, again HUN performs best, followed closely by ITA. While the good performance of HUN as a source language is still unexpected, ITA is closely related to SPA, which could explain the high accuracy of the final model. As for ENG, pretraining on EUS and NAV yields the worst final models -- importantly, accuracy is over $15\%$ lower than for QVH, which is also an unrelated language. This again suggests that the prefixing morphology of EUS and NAV might play a role. 

Lastly, for ZUL, all models perform rather poorly, with a minimum accuracy of 10.7 and 10.8 for the source languages QVH and EUS, respectively,
and a maximum accuracy of 24.9 for a model pretrained on Turkish. The latter result hints at the fact that a regular and agglutinative morphology might be beneficial in a source language -- something which could also account for the performance of models pretrained on HUN. 

\begin{table}[t] 
  \setlength{\tabcolsep}{2.5pt}
  \small
  \centering
  \begin{tabular}{l | l l l l l l l l}
  \toprule
    & EUS & FRA & DEU & HUN & ITA & NAV & TUR & QVH \\ \midrule
    ENG &44.2 & 75.8 & 81.4 & 84.5 & 84.3 & 50.8 & 81.6 & 67.3 \\ 
    SPA & 24.5 & 55.1 & 54.8 & 61.0 & 58.3 & 33.6 & 51.9 & 51.8 \\ 
 ZUL & 12.4 & 21.8 & 24.5 & 25.7 & 22.2 & 13.8 & 28.7 & 12.2 \\
  \bottomrule
  \end{tabular}
  \caption{\label{tab:acc_val} Validation accuracy.} 
\end{table}

\section{Qualitative Results}
For our qualitative analysis, we make use of the validation set. Therefore, we show validation set accuracies in Table \ref{tab:acc_val} for comparison. As we can see, the results are similar to the test set results for all language combinations.
We manually annotate the outputs for the first $75$ development examples for each source--target language combination. All found errors are categorized as belonging to one of the following categories.

\paragraph{Stem Errors}
\begin{itemize}
    
    \item \textbf{SUB(X)}: This error consists of a wrong substitution of one character with another. SUB(V) and SUB(C) denote this happening with a vowel or a consonant, respectively. Letters that differ from each other by an accent count as different vowels. 
    \newline \textbf{Example}: \textit{decultared} instead of \textit{decultured}
    
    \item \textbf{DEL(X)}: This happens when the system ommits a letter from the output. DEL(V) and DEL(C) refer to a missing vowel or consonant, respectively.
    \newline \textbf{Example}: \textit{firte} instead of \textit{firtle}
    
    \item \textbf{NO\_CHG(X)}: This error occurs when inflecting the lemma to the gold form requires a change of either a vowel (NO\_CHG(V)) or a consonant (NO\_CHG(C)), but this is missing in the predicted form.
    \newline \textbf{Example}: \textit{verto} instead of \textit{vierto}
    
    \item \textbf{MULT}: This describes cases where two or more errors occur in the stem. Errors concerning the affix are counted for separately. 
    \newline \textbf{Example}: \textit{aconcoonaste} instead of \textit{acondicionaste}
    
    \item \textbf{ADD(X)}: This error occurs when a letter is mistakenly added to the inflected form. ADD(V) refers to an unnecessary vowel, ADD(C) refers to an unnecessary consonant.
    \newline \textbf{Example}: \textit{compillan} instead of \textit{compilan}
    
    \item \textbf{CHG2E(X)}: This error occurs when inflecting the lemma to the gold form requires a change of either a vowel (CHG2E(V)) or a consonant (CHG2E(C)), and this is done, but the resulting vowel or consonant is incorrect.
    \newline \textbf{Example}: \textit{propace} instead of \textit{propague}
\end{itemize}

\paragraph{Affix Errors}
\begin{itemize}
    \item \textbf{AFF}: This error refers to a wrong affix. This can be either a prefix or a suffix, depending on the correct target form.
    \newline \textbf{Example}: \textit{ezoJulayi} instead of \textit{esikaJulayi}
    
    \item \textbf{CUT}: This consists of cutting too much of the lemma's prefix or suffix before attaching
    the inflected form's prefix or suffix, respectively.
    \newline \textbf{Example}: \textit{irradiseis} instead of \textit{irradiaseis}
\end{itemize}

\paragraph{Miscellaneous Errors}
\begin{itemize}
    \item \textbf{REFL}: This happens when a reflective pronoun is missing in the generated form.
    \newline \textbf{Example}: \textit{doli\'{e}ramos} instead of \textit{nos doli\'{e}ramos}
    
    \item \textbf{REFL\_LOC}: This error occurs if the reflective pronouns appears at an unexpected position within the generated form.
    \newline \textbf{Example}: \textit{taparsebais} instead of \textit{os tapabais}
    
    \item \textbf{OVERREG}: Overregularization errors occur when the model predicts a form which would be correct if the lemma's inflections were regular but they are not.
    \newline \textbf{Example}: \textit{underteach} instead of \textit{undertaught}
\end{itemize}

\subsection{Error Analysis: English}
\begin{table}[t] 
  \setlength{\tabcolsep}{1.2pt}
  \small
  \centering
  \begin{tabular}{l | r r r r r r r r}
  \toprule
& EUS & FRA & DEU & HUN & ITA & NAV & QVH & TUR \\ \midrule \midrule
SUB(V) & 2 & 2 & 0 & 2 & 2 & 2 & 0 & 3 \\
DEL(C) & 5 & 2 & 1 & 1 & 1 & 8 & 2 & 1 \\
DEL(V) & 6 & 1 & 2 & 0 & 2 & 5 & 4 & 1 \\
NO\_CHG(V) & 1 & 1 & 0 & 1 & 1 & 2 & 3 & 1 \\
MULT & 18 & 3 & 3 & 0 & 1 & 13 & 13 & 0 \\
ADD(V) & 0 & 0 & 0 & 0 & 0 & 2 & 0 & 0 \\
CHG2E(V) & 0 & 0 & 0 & 0 & 0 & 0 & 0 & 0 \\
ADD(C) & 5 & 0 & 0 & 0 & 0 & 3 & 0 & 0 \\
CHG2E(C) & 0 & 0 & 0 & 0 & 0 & 0 & 0 & 0 \\
NO\_CHG(C) & 0 & 0 & 0 & 0 & 0 & 0 & 0 & 0 \\
\midrule
AFF & 10 & 8 & 3 & 5 & 5 & 9 & 9 & 8 \\
CUT & 0 & 0 & 1 & 0 & 0 & 0 & 0 & 0 \\
\midrule
REFL & 0 & 0 & 0 & 0 & 0 & 0 & 0 & 0 \\
REFL\_LOC & 0 & 0 & 0 & 0 & 0 & 0 & 0 & 0 \\
OVERREG & 1 & 1 & 1 & 1 & 1 & 1 & 1 & 1 \\
\midrule \midrule
Stem & 37 & 9 & 6 & 4 & 7 & 35 & 22 & 6 \\
Affix & 10 & 8 & 4 & 5 & 5 & 9 & 9 & 8 \\
Misc & 1 & 1 & 1 & 1 & 1 & 1 & 1 & 1 \\
  \bottomrule
  \end{tabular}
  \caption{\label{tab:err_en} Error analysis for ENG as the model's L2.} 
\end{table}
Table \ref{tab:err_en} displays the errors found in the 75 first ENG development examples, for each source language.  From Table \ref{tab:acc_val}, we know that HUN $>$ ITA $>$ TUR $>$ DEU $>$ FRA $>$ QVH $>$ NAV $>$ EUS, and we get a similar picture when analyzing the first examples. Thus, especially keeping HUN and TUR in mind, we cautiously propose a first conclusion: \textit{familiarity with languages which exhibit an agglutinative morphology simplifies learning of a new language's morphology}.

Looking at the types of errors, we find that EUS and NAV make the most stem errors. For QVH we find less, but still over 10 more than for the remaining languages. This makes it seem that models pretrained on prefixing or partly prefixing languages indeed have a harder time to learn ENG inflectional morphology, and, in particular, to copy the stem correctly. Thus, our second hypotheses is that \textit{familiarity with a prefixing language might lead to suspicion of needed changes to the part of the stem which should remain unaltered in a suffixing language}. DEL(X) and ADD(X) errors are particularly frequent for EUS and NAV, which further suggests this conclusion.  

Next, the relatively large amount of stem errors for QVH leads to our second hypothesis: \textit{language relatedness does play a role when trying to produce a correct stem of an inflected form}. This is also implied by the number of MULT errors for EUS, NAV and QVH, as compared to the other languages.

Considering errors related to the affixes which have to be generated, we find that DEU, HUN and ITA make the fewest. This further suggests the conclusion that, especially since DEU is the language which is closest related to ENG, \textit{language relatedness plays a role for producing suffixes of inflected forms} as well.

Our last observation is that many errors are not found at all in our data sample, e.g., CHG2E(X) or NO\_CHG(C). This can be explained by ENG having a relatively poor inflectional morphology, which does not leave much room for mistakes.

\subsection{Error Analysis: Spanish}
\begin{table}[t] 
  \setlength{\tabcolsep}{1.2pt}
  \small
  \centering
  \begin{tabular}{l | r r r r r r r r}
  \toprule
& EUS & FRA & DEU & HUN & ITA & NAV & QVH & TUR \\ \midrule \midrule
SUB(V) & 7 & 1 & 4 & 4 & 3 & 4 & 3 & 4 \\
DEL(C) & 4 & 0 & 0 & 0 & 0 & 1 & 1 & 0 \\
DEL(V) & 4 & 0 & 1 & 0 & 0 & 2 & 0 & 0 \\
NO\_CHG(V) & 6 & 7 & 6 & 5 & 5 & 3 & 5 & 6 \\
MULT & 8 & 2 & 0 & 0 & 0 & 9 & 0 & 2 \\
ADD(V) & 4 & 2 & 0 & 0 & 0 & 0 & 1 & 0 \\
CHG2E(V) & 1 & 0 & 0 & 1 & 0 & 1 & 1 & 0 \\
ADD(C) & 3 & 1 & 1 & 0 & 0 & 3 & 0 & 1 \\
CHG2E(C) & 0 & 0 & 1 & 0 & 0 & 1 & 0 & 0 \\
NO\_CHG(C) & 0 & 0 & 0 & 0 & 1 & 0 & 1 & 0 \\
\midrule
AFF & 35 & 29 & 27 & 23 & 26 & 35 & 31 & 30 \\
CUT & 9 & 1 & 2 & 1 & 1 & 8 & 3 & 1 \\
\midrule
REFL & 2 & 0 & 2 & 0 & 1 & 2 & 1 & 1 \\
REFL\_LOC & 0 & 2 & 0 & 2 & 1 & 0 & 1 & 1 \\
OVERREG & 0 & 0 & 0 & 0 & 0 & 0 & 0 & 0 \\
\midrule \midrule
Stem & 37 & 13 & 13 & 10 & 9 & 24 & 12 & 13 \\
Affix & 44 & 30 & 29 & 24 & 27 & 43 & 34 & 31 \\
Misc & 2 & 2 & 2 & 2 & 2 & 2 & 2 & 2 \\
  \bottomrule
  \end{tabular}
  \caption{\label{tab:err_es} Error analysis for SPA as the model's L2.} 
\end{table}
The errors committed for SPA are shown in Table \ref{tab:err_es}, again listed by source language. Together with Table \ref{tab:acc_val} it gets clear that SPA inflectional morphology is more complex than that of ENG: systems for all source languages perform worse.

Similarly to ENG, however, we find that most stem errors happen for the source languages EUS and NAV, which is further evidence for our previous hypothesis that \textit{familiarity with prefixing languages impedes acquisition of a suffixing one}. Especially MULT errors are much more frequent for EUS and NAV than for all other languages. ADD(X) happens a lot for EUS, while ADD(C) is also frequent for NAV. Models pretrained on either language have difficulties with vowel changes, which reflects in NO\_CHG(V). Thus, we conclude that this phenomenon is generally hard to learn. 

Analyzing next the errors concerning affixes, we find that models pretrained on HUN, ITA, DEU, and FRA (in that order) commit the fewest errors. This supports two of our previous hypotheses: First, given that ITA and FRA are both from the same language family as SPA, \textit{relatedness seems to be benficial for learning of the second language}. Second, the system pretrained on HUN performing well suggests again that \textit{a source language with an agglutinative, as opposed to a fusional, morphology seems to be beneficial} as well.

\subsection{Error Analysis: Zulu}
\begin{table}[t] 
  \setlength{\tabcolsep}{1.2pt}
  \small
  \centering
  \begin{tabular}{l | r r r r r r r r}
  \toprule
& EUS & FRA & DEU & HUN & ITA & NAV & QVH & TUR \\ \midrule \midrule
SUB(V) & 3 & 2 & 1 & 3 & 0 & 6 & 7 & 1 \\
DEL(C) & 4 & 6 & 1 & 4 & 6 & 3 & 2 & 2 \\
DEL(V) & 1 & 7 & 0 & 2 & 2 & 0 & 3 & 1 \\
NO\_CHG(V) & 2 & 0 & 0 & 0 & 0 & 1 & 1 & 0 \\
MULT & 30 & 8 & 13 & 10 & 11 & 21 & 31 & 9 \\
ADD(V) & 0 & 1 & 1 & 3 & 1 & 2 & 0 & 2 \\
CHG2E(V) & 0 & 0 & 0 & 0 & 0 & 0 & 0 & 0 \\
ADD(C) & 1 & 3 & 1 & 6 & 4 & 2 & 1 & 1 \\
CHG2E(C) & 0 & 0 & 0 & 0 & 0 & 0 & 0 & 0 \\
NO\_CHG(C) & 0 & 2 & 1 & 1 & 1 & 0 & 0 & 1 \\
\midrule
AFF & 59 & 52 & 52 & 53 & 53 & 55 & 57 & 52 \\
CUT & 1 & 3 & 2 & 5 & 3 & 2 & 3 & 4 \\
\midrule
REFL & 0 & 0 & 0 & 0 & 0 & 0 & 0 & 0 \\
REFL\_LOC & 0 & 0 & 0 & 0 & 0 & 0 & 0 & 0 \\
OVERREG & 0 & 0 & 0 & 0 & 0 & 0 & 0 & 0 \\
\midrule \midrule
Stem & 41 & 29 & 18 & 29 & 25 & 35 & 45 & 17 \\
Affix & 60 & 55 & 54 & 58 & 56 & 57 & 60 & 56 \\
Misc & 0 & 0 & 0 & 0 & 0 & 0 & 0 & 0 \\
  \bottomrule
  \end{tabular}
  \caption{\label{tab:err_zul} Error analysis for ZUL as the model's L2.} 
\end{table}
In Table \ref{tab:err_zul}, the errors for Zulu are shown, and Table \ref{tab:acc_val} reveals the relative performance for different source languages: TUR $>$ HUN $>$ DEU $>$ ITA $>$ FRA $>$ NAV $>$ EUS $>$ QVH.
Again, TUR and HUN obtain high accuracy, which is an additional indicator for our hypothesis that \textit{a source language with an agglutinative morphology facilitates learning of inflection in another language}.

Besides that, results differ from those for ENG and SPA. First of all, more mistakes are made for all source languages. However, there are also several finer differences. For ZUL, the model pretrained on QVH makes the most stem errors, in particular 4 more than the EUS model, which comes second. Given that ZUL is a prefixing language and QVH is suffixing, this relative order seems important. QVH also committs the highest number of MULT errors. 

The next big difference between the results for ZUL and those for ENG and SPA is that DEL(X) and ADD(X) errors, which previously have mostly been found for the prefixing or partially prefixing languages EUS and NAV, are now most present in the outputs of \textit{suffixing} languages. Namely, DEL(C) occurs most for FRA and ITA, DEL(V) for FRA and QVH, and ADD(C) and ADD(V) for HUN. While some deletion and insertion errors are subsumed in MULT, this does not fully explain this difference. For instance, QVH has both the second most DEL(V) and the most MULT errors.

The overall number of errors related to the affix seems comparable between models with different source languages. This weakly supports the hypothesis that \textit{relatedness reduces affix-related errors}, since none of the pretraining languages in our experiments is particularly close to ZUL. However, we do find more CUT errors for HUN and TUR: again, these are suffixing, while CUT for the target language SPA mostly happened for the prefixing languages EUS and NAV.

\subsection{Limitations}
A limitation of our work is that we only include languages that are written in Latin script. 
An interesting question for future work might, thus, regard the effect of disjoint L1 and L2 alphabets.

Furthermore, none of the languages included in our study exhibits a templatic morphology. We make this choice because data for templatic languages is currently mostly available in non-Latin alphabets. Future work could investigate languages with templatic morphology as source or target languages, if needed by mapping the language's alphabet to Latin characters.

Finally, while we intend to choose a diverse set of languages for this study, our overall number of languages is still rather small. This affects the generalizability of the results, and future work might want to look at larger samples of languages.

\section{Related Work}
\paragraph{Neural network models for inflection. } Most research on inflectional morphology in NLP within the last years has been related to the SIGMORPHON and CoNLL--SIGMORPHON shared tasks on morphological inflection, which have been organized yearly since 2016 \cite{W16-2002}. Traditionally being focused on individual languages, the 2019 edition \cite{mccarthy-etal-2019-sigmorphon} contained a task which asked for transfer learning from a high-resource to a low-resource language. However, source--target pairs were predefined, and the question of how the source language influences learning besides the final accuracy score was not considered. 
Similarly to us, \newcite{kyle} performed a manual error analysis of morphological inflection systems for multiple languages. However, they did not investigate transfer learning, but focused on monolingual models.

Outside the scope of the shared tasks, \newcite{kann-etal-2017-one} investigated cross-lingual transfer for morphological inflection, but was limited to a quantitative analysis. Furthermore, that work experimented with a standard sequence-to-sequence model \cite{bahdanau2015neural} in a multi-task training fashion \cite{caruana1997multitask}, while we pretrain and fine-tune pointer--generator networks. 
 \newcite{jin-kann-2017-exploring} 
 also investigated cross-lingual transfer in neural sequence-to-sequence models for morphological inflection. However, their experimental setup mimicked \newcite{kann-etal-2017-one}, and the main research questions were different: While \newcite{jin-kann-2017-exploring} asked how cross-lingual knowledge transfer works during multi-task training of neural sequence-to-sequence models on two languages, we investigate if  neural inflection models demonstrate interesting differences in production errors depending on the pretraining language. 
Besides that, we differ in the artificial neural network architecture and language pairs we investigate.

\paragraph{Cross-lingual transfer in NLP.}
Cross-lingual transfer learning has been used for a large variety NLP of
tasks, e.g., automatic speech recognition
\cite{huang2013cross}, entity
recognition \cite{MengqiuWang2014}, language modeling \cite{tsvetkov-EtAl:2016:N16-1}, or parsing
\cite{CohenDS11,sogaard:2011:ACL-HLT20112,TACL892}. Machine translation has been no exception \cite{zoph-knight:2016:N16-1,ha2016toward,TACL1081}. 
Recent research asked how to automatically select a suitable source language for a given target language \cite{lin-etal-2019-choosing}. This is similar to our work in that our findings could potentially be leveraged to find good source languages. 

\paragraph{Acquisition of morphological inflection.}
Finally, a lot of research has focused on human L1 and L2 acquisition of inflectional morphology \cite{salaberry2000acquisition,herschensohn2001missing,housen2002corpus,ionin2002easier,weerman2006l1,zhang2010acquisition}. 

To name some specific examples, \newcite{marques2011study} investigated the effect of a stay abroad on Spanish L2 acquisition, including learning of its verbal morphology in English speakers.
\newcite{jia2003acquisition} studied how Mandarin Chinese-speaking children learned the English plural morpheme.
\newcite{nicoladis2012young} studied the English past tense acquisition in Chinese--English and French--English bilingual children. They found that, while both groups showed similar production accuracy, they differed slightly in the type of errors they made. Also considering the effect of the native language explicitly, \newcite{yang2004impact} investigated the acquisition of the tense-aspect system in an L2 for speakers of a native language which does not mark tense explicitly.

Finally, our work has been weakly motivated by \newcite{bliss2006l2}. There, the author asked a  question for human subjects which is similar to the one we ask for neural models: How does the native language influence L2 acquisition of inflectional morphology?

\section{Conclusion and Future Work}
Motivated by the fact that, in humans, learning of a second language is influenced by a learner's native language, we investigated a similar question in artificial neural network models for morphological inflection: How does pretraining on different languages influence a model's learning of inflection in a target language?

We performed experiments on eight different source languages and three different target languages. An extensive error analysis of all final models showed that (i) for closely related source and target languages, acquisition of target language inflection gets easier; (ii) knowledge of a prefixing language makes learning of inflection in a suffixing language more challenging, as well as the other way around; and (iii) languages which exhibit an agglutinative morphology facilitate learning of inflection in a second language.

Future work might leverage those findings to improve neural network models for morphological inflection in low-resource languages, by choosing suitable source languages for pretraining. 

Another interesting next step would be to investigate how the errors made by our models compare to those by human L2 learners with different native languages. If the exhibited patterns resemble each other, computational models could be used to predict errors a person will make, which, in turn, could be leveraged for further research or the development of educational material.

\section*{Acknowledgments}
I would like to thank Samuel R. Bowman and Kyle Gorman for helpful discussions and suggestions. 
This work has benefited from the support of Samsung Research under the project \textit{Improving Deep Learning using Latent Structure} and from the donation of a Titan V GPU by NVIDIA Corporation.

\bibliography{naaclhlt2019}
\bibliographystyle{acl_natbib}

\appendix

\end{document}